\documentclass[english]{lni}
\usepackage{booktabs}
\usepackage{xcolor}
\usepackage{amsmath}
\usepackage{graphicx}
\usepackage{multirow}
\usepackage{float}

\begin{document}
\title[Grokking]{Is Grokking a Loss of Normal Hyperbolicity of the Interpolation Manifold?}

\author[1]{Suvinava Basak}{suvinava.basak@tu-braunschweig.de}{0009-0005-4289-1205}
\affil[1]{Technische Universit\"at Braunschweig\\Data Science\\ Universit\"atsplatz 2\\38106 Braunschweig\\Germany}

\newcommand{\sminp}{\sigma_{\min}^{+}}
\newcommand{\R}{\mathbb{R}}
\newcommand{\symbolfootnote}[1]{%
  \begingroup
  \renewcommand{\thefootnote}{\fnsymbol{footnote}}%
  \setcounter{footnote}{0}%
  \stepcounter{footnote}%
  \footnotetext{#1}%
  \endgroup
}

\maketitle
\symbolfootnote{Accepted for publication at SKILL 2026 (GI-Edition Lecture Notes in Informatics, Gesellschaft für Informatik e.V.).}

\begin{abstract}
A recent line of work recasts the post-memorization phase of grokking as constrained optimization: once a network interpolates the training set, weight decay drives a slow drift along the zero-loss manifold toward lower norm. In the language of dynamical systems, this is a fast--slow system in which the interpolation manifold plays the role of a slow manifold. We ask a question that this framing makes natural but the existing literature does not address: \emph{is the sharp generalization transition a loss of normal hyperbolicity of that manifold}: a fold- or bifurcation-like event in which a normal restoring direction goes flat? Or does the manifold stay uniformly attracting while generalization happens by smooth drift? We propose a simple, optimizer-agnostic diagnostic: the smallest nonzero singular value $\sminp(\mathbf J)$ of the residual Jacobian, which, for the squared loss, equals the slowest normal restoring rate of the manifold. On a two-layer ReLU network trained to
grok modular addition under squared loss, $\sminp(\mathbf J)$ does not collapse at the transition; it is near zero only \emph{before} memorization and attains its largest values \emph{during} the transition. The result holds across five seeds, and the six smallest singular values behave identically; there is no subspace-local collapse either. This is preliminary evidence against the bifurcation hypothesis and in favor of the smooth-contraction picture. We are explicit that a single-setting, gradual-transition experiment under \texttt{Adam} optimizer does not prove the absence of a bifurcation; it constrains where one could hide.
\end{abstract}


\section{Introduction}
\label{sec:intro}

Grokking is the empirical observation that a network can reach perfect training accuracy with near-chance test accuracy and only generalize after a long additional period of training~\cite{power2022}. It is often discussed alongside \emph{double descent}, in which test error first falls, then rises as the model overfits, and eventually falls again~\cite{belkin2019reconciling, nakkiran2021deep}. The two are related but not identical: Power et al.~\cite{power2022} distinguished grokking on the grounds that its generalization occurs far past the interpolation threshold, while Davies et al.~\cite{davies2023unifying} argue both arise from a single pattern-learning-speed mechanism. Our question is orthogonal to that taxonomy: we ask about the \emph{geometry} of the grokking transition rather than its classification.

A productive recent account treats the long post-memorization phase as constrained optimization. Mu\c{s}at~\cite{musat2025} argues that, after interpolation, gradient descent with small weight decay effectively minimizes the parameter norm subject to staying on the zero-loss manifold, in the limit of infinitesimal learning rate and weight decay; a concurrent framework reaches the same endpoint through a Riemannian norm flow on the manifold of interpolating solutions~\cite{boursier2025}. Lyu et al.~\citet{lyu2024} prove, for homogeneous networks with large initialization and small weight decay, a sharp transition from a kernel predictor to a KKT point of a global min-norm/max-margin problem~\cite{soudry2018,lyuli2020} on both classification and regression, with modular addition under cross-entropy as the motivating phenomenon.

All of these share a structure: a fast process pulls the parameters onto a low-dimensional set of (near-)interpolating solutions, and a slow process driven by weight decay moves the parameters \emph{along} that set. This is exactly a singularly perturbed, or fast-slow, dynamical system, with the interpolation manifold as the slow manifold. Yet none of these works invokes the corresponding mathematics: normal hyperbolicity, Fenichel persistence of slow manifolds~\cite{fenichel1979, kuehn2015}, and what happens when normal hyperbolicity is \emph{lost}. We ask whether grokking's defining feature,
the suddenness of the transition, is a signature of that loss.
 
Concretely, we contrast two hypotheses. \textbf{(H1)} The interpolation manifold loses normal hyperbolicity at the transition: a normal restoring direction softens toward zero, the fast/slow timescale separation breaks, and the trajectory jumps: a fold or pitchfork in the fast--slow
decomposition. Modular addition's symmetry group makes a symmetry-breaking pitchfork a natural candidate normal form.
\textbf{(H0)} The manifold stays uniformly normally hyperbolic; the transition is a feature of the slow drift through a region where test behavior changes quickly, without any degeneration of the manifold's geometry. This is the picture implied by the contraction-based accounts above. If \textbf{H1} holds true, this would tie the dynamical and mechanistic-interpretability~\cite{nanda2023} views together. Our contribution is a clean test that distinguishes H1 from H0, and a preliminary, and as it turns out, negative result.

\section{The diagnostic}
\label{sec:diag}

Let $f(\boldsymbol \theta;\mathbf x)\in\R^{c}$ be the network output and $L(\boldsymbol \theta)=\frac1n\sum_{i=1}^{n}\lVert f(\boldsymbol \theta;\mathbf x_i)- \mathbf y_i\rVert^2$ the squared loss on a training set of size $n$. We use squared loss precisely because it makes the \emph{interpolation manifold} $\mathcal{M}=\{\boldsymbol \theta : f(\boldsymbol \theta;\mathbf x_i)= \mathbf y_i\ \forall i\}$ an exact object with a well-defined tangent and normal structure. Training with decoupled weight decay $\lambda$ follows, in the gradient-flow limit, $\dot{\boldsymbol \theta}=-\nabla L(\boldsymbol \theta)-\lambda \boldsymbol \theta$; for small $\lambda$ the term $-\nabla L$ is fast and pulls toward $\mathcal{M}$, while $-\lambda \boldsymbol \theta$ is the slow drift along it.

Collect residuals into $\mathbf r(\boldsymbol \theta)=(f(\boldsymbol \theta;\mathbf x_i)- \mathbf y_i)_i\in\R^{nc}$ and let $\mathbf J(\boldsymbol \theta)=\partial r/\partial \boldsymbol \theta\in\R^{nc\times D}$ be the residual Jacobian. The Gauss--Newton matrix is $\mathbf G=\mathbf J^\top \mathbf J$. At a point of $\mathcal{M}$ the row space of $\mathbf J$ is the normal space $N_\theta\mathcal{M}$ (directions in which residuals, hence the loss, change to first order), and $\ker \mathbf J$ is the tangent space $T_\theta\mathcal{M}$ (directions along which the model stays interpolating). The nonzero eigenvalues of $\mathbf G$ are the squared
singular values $\sigma_i(\mathbf J)^2$, so the smallest normal restoring rate is
\begin{equation*}
  \hspace{4.5cm} \kappa(\boldsymbol \theta)=\sminp\!\big(\mathbf J(\boldsymbol \theta)\big)^2,
  \label{eq:kappa}
\end{equation*}
where $\sminp$ is the smallest nonzero singular value. For an over-parameterized network with $nc<D$, $\mathbf J$ has generically full row rank, so all $nc$ singular values are positive and $\sminp$ is their minimum.
 
\paragraph{Why $\sminp$ captures normal contraction.}
Near an interpolating point $\boldsymbol{\theta}^\star$ (so $\mathbf r(\boldsymbol \theta^\star)=0$) the fast flow is $\dot{\boldsymbol \theta}=-\nabla L(\boldsymbol \theta)$ with $\nabla L=\tfrac{2}{n} \mathbf J^\top \mathbf r$. Its linearization is governed by the Hessian $\nabla^2 L=\tfrac{2}{n}\big(\mathbf J^\top \mathbf J+\sum_k r_k\nabla^2 r_k\big)$, whose second term \emph{vanishes at $r=0$}; hence $\nabla^2 L(\boldsymbol \theta^\star)=\tfrac{2}{n}\mathbf J^\top \mathbf J$, the Gauss--Newton matrix. Writing $\partial \boldsymbol \theta= \partial \boldsymbol \theta_\parallel + \partial \boldsymbol \theta_\perp$ along $T_{\boldsymbol\theta^\star}\mathcal{M}=\ker \mathbf J$ and $N_{\boldsymbol \theta^\star}\mathcal{M}=\mathrm{row}(\mathbf J)$, the linearized fast dynamics is $\dot{\partial \boldsymbol \theta}\approx-\tfrac{2}{n}\mathbf J^\top \mathbf J\,\partial \boldsymbol \theta$: tangential components lie in $\ker \mathbf J$ and are not restored (the slow directions), while normal components contract at the rates $\tfrac{2}{n}\sigma_i(\mathbf J)^2$. The slowest normal contraction rate is therefore $\propto\sminp(\mathbf J)^2$. Normal hyperbolicity of $\mathcal{M}$ is exactly the statement that this rate is bounded away from zero uniformly along the slow
drift -- the condition under which Fenichel theory guarantees the perturbed slow manifold persists and the reduced flow is the norm flow of~\cite{musat2025,boursier2025}. A loss of normal hyperbolicity, the
prerequisite for the fold/pitchfork of H1, is exactly $\sminp(\mathbf J)\to 0$. The diagnostic is thus the single scalar trajectory $\sminp(\mathbf J(\boldsymbol \theta(t)))$ read against test accuracy: a dip toward zero at the transition supports H1; a value bounded away from zero supports H0. Since $\sminp(\mathbf J)$ depends only on the model and data at $\theta$, it is well defined under any optimizer, not only the flow for which the slow-drift theory is cleanest.

\section{Experimental setup}
\label{sec:setup}

We train a two-layer ReLU network $f(\boldsymbol \theta;\mathbf x)=\mathbf W_2\,\mathrm{ReLU}(\mathbf W_1 \mathbf x+\mathbf b_1)$ with width $96$ and no second-layer bias on modular addition mod $p=11$: the input concatenates the one-hot encodings of $a$ and $b$, and the target is the one-hot encoding of $(a+b)\bmod p$. We use a random $70\%/30\%$ train/test split, squared loss,
full-batch optimization, and a large initialization (Kaiming $\times 3.5$) to induce a clear memorization plateau~\cite{omnigrok}. To reach the transition within a practical compute budget, we optimize with AdamW (learning rate $3\times10^{-3}$, decoupled weight decay $2.0$) for $35{,}000$ steps. We snapshot $\boldsymbol \theta$ every $500$ steps and compute $\sminp(\mathbf J)$ (via the SVD of $\mathbf J$, in $\texttt{float32}$) in a separate pass, so the decomposition does not slow training. Because ReLU makes $\mathbf J$ piecewise constant in the activation pattern, $\mathbf J$ can change across activation switches; in practice the resulting $\sminp$ trajectory is smooth at snapshot resolution. We repeat the run over five initialization seeds.
 
\section{Results}
\label{sec:results}

\begin{figure}[t]
  \centering
  \includegraphics[width=0.80\linewidth]{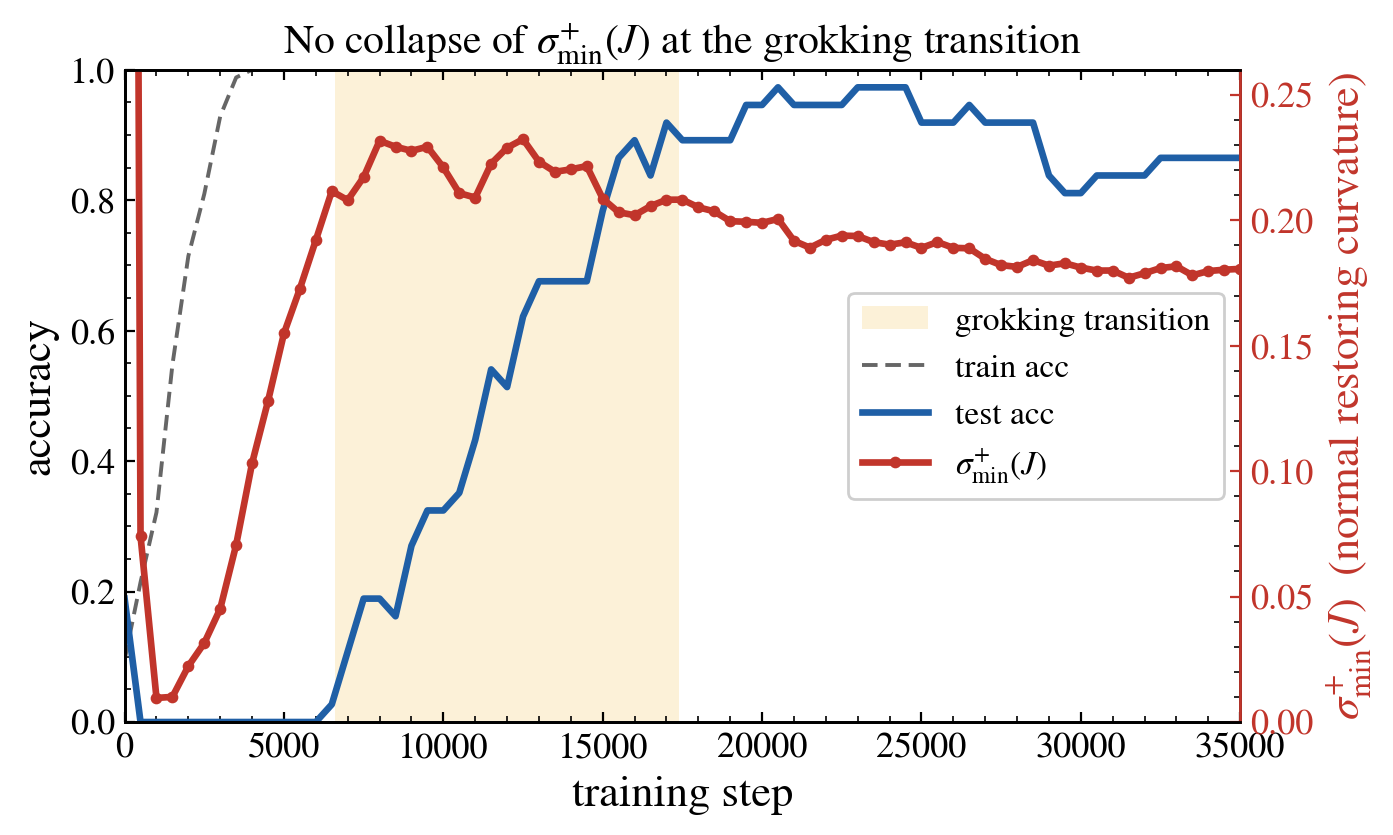}
  \caption{Test/training accuracy (left axis) and the normal restoring rate $\sminp(\mathbf J)$ (right axis). The shaded band marks the generalization transition. $\sminp(\mathbf J)$ does not dip during the transition; it is small only before memorization and largest while the model generalizes. Two-layer ReLU network, squared loss, AdamW, modular addition $p=11$; single seed.}
  \label{fig:main}
\end{figure}

The run groks (Fig.~\ref{fig:main}): training accuracy reaches $1.0$ by about step $4{,}000$; test accuracy sits at $0.0$, \emph{below} the $\approx0.09$ chance level for $11$ classes, i.e. the memorizing solution actively mispredicts test points, through about step $6{,}000$, then rises over roughly steps $7{,}000$--$17{,}000$ to $\approx0.95$ before settling near $0.86$.

The diagnostic does not behave as \textbf{H1} predicts. $\sminp(\mathbf J)$ is near zero only \emph{before} memorization: its global minimum, $\approx1\times10^{-2}$, occurs at step $1{,}000$, while training accuracy is still well below $1$ and the network has not yet differentiated its outputs. It then rises monotonically as the manifold forms, and across the entire transition it sits at its largest values; the normal restoring rate \emph{peaks} mid-transition rather than collapsing. Table~\ref{tab:phases} summarizes the medians by phase.

\begin{table}[t]
  \centering
  \begin{tabular}{lccc}
    \toprule \toprule
    \textbf{Phase} & \textbf{Step range} & \textbf{median} $\sminp(\mathbf J)$ & \textbf{min} $\sminp(\mathbf J)$\\
    \midrule
    Pre-memorization & $500$--$1800$ & $\approx0.010$ & $0.0096$\\
    Memorization plateau & $3000$--$6000$ & $0.128$ & $0.045$\\
    Generalization transition & $7000$--$17000$ & $0.220$ & $0.202$\\
    Post-transition & $\ge22000$ & $0.182$ & $0.177$\\
    \bottomrule
  \end{tabular}
  \caption{Normal restoring rate by training phase (single seed). The only regime in which $\sminp(\mathbf J)$ approaches zero is pre-memorization, which is unrelated to grokking; during the transition it is at its maximum. The pre-memorization reading also shows the diagnostic is sensitive enough to register a genuinely near-degenerate Jacobian when one occurs.}
  \label{tab:phases}
\end{table}

\begin{figure}[t]
  \centering
  \includegraphics[width=\linewidth]{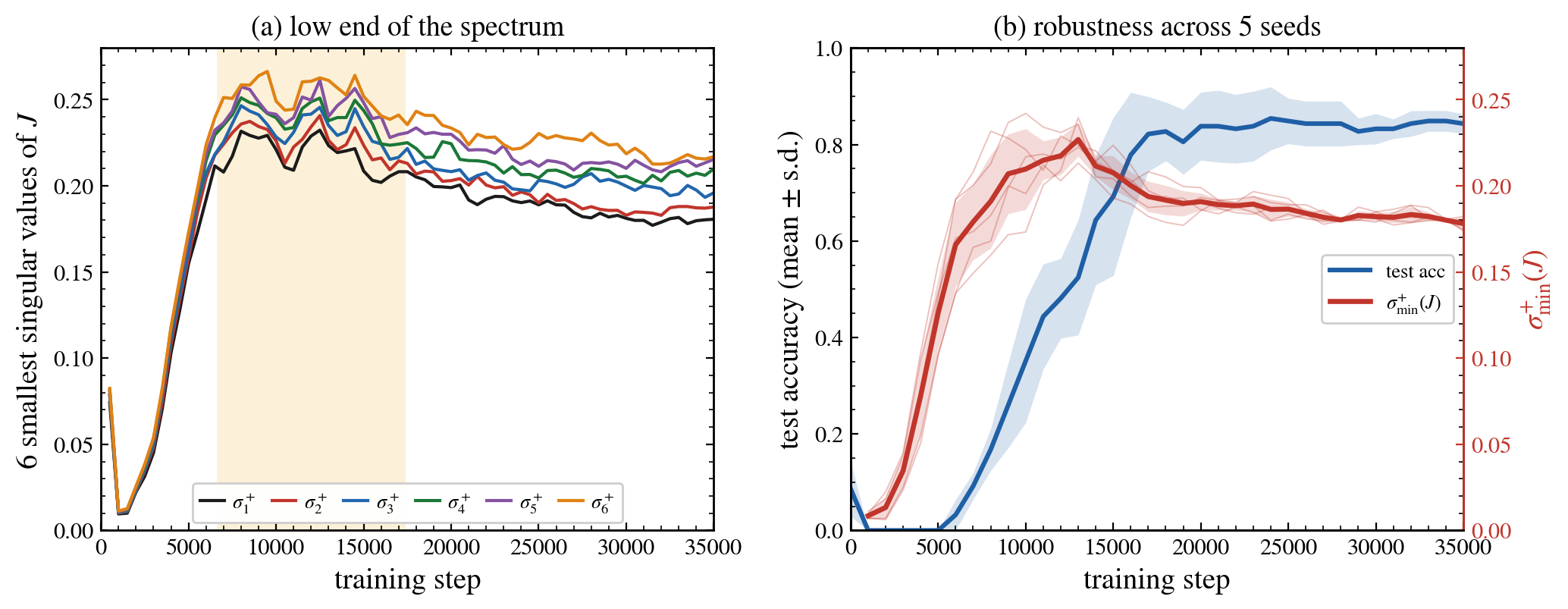}
  \caption{(a) The six smallest singular values ($\sigma_1^+$ being the smallest) of $\mathbf J$ form a tight cluster bounded away from zero across the transition; the smallest dips only pre-memorization. The yellow shaded region identifies the grokking transition. (b) Robustness across five seeds: thin red lines are per-seed $\sminp(\mathbf J)$, the bold red curve and band are mean $\pm$ s.d., and the blue curve/band is mean $\pm$ s.d.\ test accuracy. No seed shows a dip at its transition.}
  \label{fig:checks}
\end{figure}

Two checks address the natural objections that a single number or a single seed could mislead (Fig.~\ref{fig:checks}). First, the \emph{six} smallest singular values behave identically: they form a tight cluster that stays bounded away from zero throughout the transition, so the absence of
a dip is not an artifact of looking only at the global minimum, no low-dimensional subspace collapses either (panel a). Second, the result is robust across five seeds: although the transition occurs at different steps per seed (test accuracy first exceeds $0.5$ between steps $\approx10{,}000$
and $16{,}000$), in every seed $\sminp(\mathbf J)$ during its transition window remains in the $0.18$--$0.23$ range, and the seed-averaged curve shows no dip (panel b).

A secondary observation: the parameter norm does not decrease through the transition in this run (it stays near $10$ and drifts slightly upward), so generalization is not accompanied by visible $\ell_2$-norm contraction here. This is consistent with AdamW's implicit bias being governed by an
$\ell_\infty$-type rather than $\ell_2$ geometry~\cite{xie2024,
gronich2026}, and we treat it as a caveat about the optimizer rather than a claim about the manifold. We refrain from reporting the dynamical
timescale ratio $\sminp(\mathbf J)^2/\lambda$: for the gradient-flow theory $\lambda$ is the drift rate, but AdamW's decoupled weight decay does not play that role (Adam rescales the effective step per coordinate), so the raw ratio would misstate the separation; a clean ratio belongs to a gradient-flow study.

\section{Discussion and limitations}
Taken at face value the experiment favors \textbf{H0} over \textbf{H1}: the interpolation manifold remains robustly normally hyperbolic across grokking, and the suddenness of the transition is not mirrored by any softening of a normal restoring direction. This is the outcome expected from the smooth-contraction picture of~\cite{musat2025,boursier2025,lyu2024} and from recent work framing the grokking delay as exponential norm contraction across a geometric gap: Truong et al.~\cite{delaylaw} derive a norm-separation delay
law\footnote{We cite it for context and have not independently reproduced it.} for discrete regularized SGD. Because the entry--exit delay law of a delayed-loss-of-stability mechanism only applies if a bifurcation exists, the negative result also removes the motivation for that route.

We are deliberate about what this does not establish. First, we optimize with AdamW, whereas the fast--slow theory is stated for gradient descent or gradient flow; the sharpest grokking regimes (GD with large initialization and small weight decay) are untested here. Second, the observed transition is moderately gradual rather than maximally sharp. Third, $\sminp(\mathbf J)$ (and the six smallest) is still a global measure; a loss of normal hyperbolicity confined to a narrow, test-relevant subspace (for instance the symmetry-aligned directions of modular addition) could in principle occur while the low end of the spectrum stays bounded away from zero, and a subspace-resolved diagnostic would be needed to exclude it. Fourth, we measure a discrete trajectory, not the gradient-flow limit in which the manifold is exactly defined. We therefore frame the result as constraining, not refuting: in this regime, grokking is not accompanied by a global collapse of the manifold's normal restoring rate.

These limitations also name the experiments that would settle the question: the same diagnostic under gradient descent in a sharp-transition regime; a subspace-resolved curvature projected onto the directions that most separate train from test points (or onto symmetry-aligned subspaces); and, if a dip
never appears, a positive characterization of which geometric quantity \emph{does} move monotonically through the transition, which under AdamW may be an $\ell_\infty$- or sparsity-based measure rather than $\ell_2$ norm.
 
\section{Conclusion}
\label{sec:conclusion}
We asked whether the sharp transition in grokking is a loss of normal hyperbolicity of the interpolation manifold, and gave a diagnostic: the smallest nonzero singular value of the residual Jacobian that turns the question into a single measurable curve, justified for the squared loss as the slowest normal restoring rate. On a network that groks modular addition under squared loss, the answer appears to be no, robustly across seeds and across the low end of the spectrum: the manifold stays normally hyperbolic through the transition. We do not claim to have proved the absence of a bifurcation; we show that, in a clean grokking run with a sensitive instrument, no such signature is present, which is itself a constraint on theories of why the transition is sharp.
 
\paragraph{Reproducibility.} Entire training and diagnostic code, data behind Fig.~\ref{fig:main} and \ref{fig:checks}, and all the necessary scripts are available at GitHub\footnote{GitHub Repo: \url{https://github.com/baksho/grokking-normal-hyperbolicity}}.

\printbibliography

@article{power2022,
  author       = {Power, Alethea and Burda, Yuri and Edwards, Harrison and
                  Babuschkin, Igor and Misra, Vedant},
  title        = {Grokking: Generalization Beyond Overfitting on Small Algorithmic Datasets},
  journal      = {1st Mathematical Reasoning in General Artificial Intelligence Workshop},
  year         = {2022},
  eprint       = {2201.02177},
  eprinttype   = {arXiv},
  url          = {https://arxiv.org/abs/2201.02177}
}

@article{belkin2019reconciling,
  author  = {Belkin, Mikhail and Hsu, Daniel and Ma, Siyuan and Mandal, Soumik},
  title   = {Reconciling Modern Machine-Learning Practice and the Classical Bias--Variance Trade-off},
  journal = {Proceedings of the National Academy of Sciences},
  volume  = {116},
  number  = {32},
  pages   = {15849--15854},
  year    = {2019},
  doi     = {10.1073/pnas.1903070116}
}

@article{nakkiran2021deep,
  author  = {Nakkiran, Preetum and Kaplun, Gal and Bansal, Yamini and Yang, Tristan and Barak, Boaz and Sutskever, Ilya},
  title   = {Deep Double Descent: Where Bigger Models and More Data Hurt},
  journal = {Journal of Statistical Mechanics: Theory and Experiment},
  volume  = {2021},
  year    = {2021},
  doi     = {10.1088/1742-5468/ac3a74}
}

@inproceedings{davies2023unifying,
  author    = {Davies, Xander and Langosco, Lauro and Krueger, David},
  title     = {Unifying Grokking and Double Descent},
  booktitle = {NeurIPS 2022 Machine Learning Safety Workshop},
  journal   = {36th Conference on Neural Information Processing Systems},
  year      = {2022},
}

@article{musat2025,
  author      = {Musat, Tiberio},
  title       = {The Geometry of Grokking: Norm Minimization on the Zero-Loss Manifold},
  journal     = {arXiv preprint arXiv:2511.01938},
  year        = {2025},
  url         = {https://arxiv.org/abs/2511.01938},
}

@article{boursier2025,
  author        = {Etienne Boursier and Scott Pesme and
                   Radu-Alexandru Dragomir},
  title         = {A Theoretical Framework for Grokking:
                   Interpolation followed by Riemannian Norm Minimisation},
  journal       = {39th Conference on Neural Information Processing Systems (NeurIPS 2025)},
  year          = {2025},
  eprint        = {2505.20172},
  eprinttype    = {arXiv},
  url           = {https://arxiv.org/pdf/2505.20172}
}

@inproceedings{lyu2024,
  author    = {Kaifeng Lyu and Jikai Jin and Zhiyuan Li and
               Simon S. Du and Jason D. Lee and Wei Hu},
  title     = {Dichotomy of Early and Late Phase Implicit Biases
               Can Provably Induce Grokking},
  booktitle = {The 12th International Conference on Learning Representations (ICLR)},
  year      = {2024},
  eprint    = {2311.18817},
  eprinttype = {arXiv},
  url       = {https://proceedings.iclr.cc/paper_files/paper/2024/file/909c8fef63e1cede406ce9e6794f99a2-Paper-Conference.pdf}
}

@inproceedings{nanda2023,
  author    = {Neel Nanda and Lawrence Chan and Tom Lieberum and
               Jess Smith and Jacob Steinhardt},
  title     = {Progress Measures for Grokking via Mechanistic Interpretability},
  booktitle = {The 11th International Conference on Learning Representations (ICLR)},
  year      = {2023},
  eprint    = {2301.05217},
  eprinttype = {arXiv},
  url       = {https://arxiv.org/pdf/2301.05217}
}

@article{soudry2018,
  author  = {Daniel Soudry and Elad Hoffer and
             Mor Shpigel Nacson and Suriya Gunasekar and Nathan Srebro},
  title   = {The Implicit Bias of Gradient Descent on Separable Data},
  journal = {Journal of Machine Learning Research},
  volume  = {19},
  number  = {70},
  pages   = {1--57},
  year    = {2018},
  url     = {https://www.jmlr.org/papers/volume19/18-188/18-188.pdf}
}

@inproceedings{lyuli2020,
  author    = {Kaifeng Lyu and Jian Li},
  title     = {Gradient Descent Maximizes the Margin of Homogeneous Neural Networks},
  booktitle = {International Conference on Learning Representations (ICLR)},
  year      = {2020},
  eprint    = {1906.05890},
  eprinttype = {arXiv},
  url       = {https://openreview.net/pdf?id=SJeLIgBKPS}
}

@inproceedings{xie2024,
  author    = {Shuo Xie and Zhiyuan Li},
  title     = {Implicit Bias of AdamW: {$\ell_\infty$}-Norm Constrained Optimization},
  booktitle = {International Conference on Machine Learning (ICML)},
  year      = {2024},
  eprint    = {2404.04454},
  eprinttype = {arXiv},
  url       = {https://arxiv.org/pdf/2404.04454}
}

@article{gronich2026,
  author        = {Eitan Gronich and Gal Vardi},
  title         = {The Implicit Bias of Adam and Muon on Smooth Homogeneous Neural Networks},
  booktitle     = {Proceedings of the 43 rd International Conference on Machine Learning},
  year          = {2026},
  eprint        = {2602.16340},
  eprinttype    = {arXiv},
  publisher     = {PMLR},
  url           = {https://arxiv.org/pdf/2602.16340}
}

@article{fenichel1979,
  author  = {Neil Fenichel},
  title   = {Geometric Singular Perturbation Theory for Ordinary Differential Equations},
  journal = {Journal of Differential Equations},
  volume  = {31},
  number  = {1},
  pages   = {53--98},
  year    = {1979},
  doi     = {10.1016/0022-0396(79)90152-9},
  url     = {https://www.sciencedirect.com/science/article/pii/0022039679901529}
}

@book{kuehn2015,
  author    = {Christian Kuehn},
  title     = {Multiple Time Scale Dynamics},
  series     = {Applied Mathematical Sciences},
  volume     = {191},
  publisher = {Springer},
  year      = {2015},
  doi       = {10.1007/978-3-319-12316-5}
}

@inproceedings{omnigrok,
  author    = {Ziming Liu and Eric J. Michaud and Max Tegmark},
  title        = {Omnigrok: Grokking Beyond Algorithmic Data},
  booktitle    = {International Conference on Learning Representations},
  year         = {2023},
  eprint       = {2210.01117},
  eprinttype   = {arXiv},
  url          = {https://openreview.net/pdf?id=zDiHoIWa0q1},
  doi          = {10.48550/arXiv.2210.01117}
}

@article{delaylaw,
  author        = {Khanh Xuan Truong and Truong Quynh Hoa and
                   Luu Duc Trung and Phan Thanh Duc},
  title         = {The Norm-Separation Delay Law of Grokking:
                   A First-Principles Theory of Delayed Generalization},
  year          = {2026},
  eprint        = {2603.13331},
  eprinttype    = {arXiv},
  url           = {https://arxiv.org/pdf/2603.13331}
}

\end{document}